\documentclass[lettersize,journal]{IEEEtran}
\usepackage{amsmath,amsfonts}
\usepackage{algorithmic}
\usepackage{algorithm}
\usepackage{array}
\usepackage[caption=false,font=normalsize,labelfont=sf,textfont=sf]{subfig}
\usepackage{textcomp}
\usepackage{stfloats}
\usepackage{url}
\usepackage{verbatim}
\usepackage{graphicx}
\usepackage{cite}

\usepackage{times}
\usepackage{latexsym}
\usepackage{graphicx} 
\usepackage{amsmath}
\usepackage{amsfonts}
\usepackage{bm}
\usepackage{multirow}

\usepackage{algorithm}

\begin{document}

\title{Relation-dependent Contrastive Learning with Cluster Sampling for Inductive Relation Prediction}

\author{Jianfeng Wu, Sijie Mai, Haifeng Hu
\thanks{Jianfeng Wu, Sijie Mai and Haifeng Hu are with the School of Electronics and Information Technology, Sun Yat-sen University,Guangzhou, China(e-mail: wujf36@mail2.sysu.edu.cn; huhaif@mail.sysu.edu.cn; maisj@mail2.sysu.edu.cn).}% <-this % stops a space
\thanks{Haifeng Hu is the corresponding author.}}

\markboth{Journal of \LaTeX\ Class Files,~Vol.~14, No.~8, August~2021}%
{Shell \MakeLowercase{\textit{et al.}}: A Sample Article Using IEEEtran.cls for IEEE Journals}


\maketitle

\begin{abstract}
Relation prediction is a task designed for knowledge graph completion which aims to predict missing relationships between entities. Recent subgraph-based models for inductive relation prediction have received increasing attention, which can predict relation for unseen entities based on the extracted subgraph surrounding the candidate triplet. However, they are not completely inductive because of their disability of predicting unseen relations. Moreover, they fail to pay sufficient attention to the role of relation as they only depend on the model to learn parameterized relation embedding, which leads to inaccurate prediction on long-tail relations. In this paper, we introduce \textbf{Re}lation-dependent \textbf{Co}ntrastive \textbf{Le}arning (ReCoLe) for inductive relation prediction, which adapts contrastive learning with a novel sampling method based on clustering algorithm to enhance the role of relation and improve the generalization ability to unseen relations. Instead of directly learning embedding for relations, ReCoLe allocates a pre-trained GNN-based encoder to each relation to strengthen the influence of relation. The GNN-based encoder is optimized by contrastive learning, which ensures satisfactory performance on long-tail relations. In addition, the cluster sampling method equips ReCoLe with the ability to handle both unseen relations and entities. Experimental results suggest that ReCoLe outperforms state-of-the-art methods on commonly used inductive datasets. %and excels at handling long-tail and unseen relations.
\end{abstract}


\section{Introduction}
\IEEEPARstart{K}{nowledge} graphs (KGs) store factual information represented as relational triplet $(h, r, t)$ which means there is a relation $r$ between head entity $h$ and tail entity $t$. With rapid development of KGs, various KGs applications such as recommendation\cite{recommendation}, semantic search\cite{semantic_research} and question answering\cite{question_answering} have come to the fore in recent years. Nevertheless, KGs are vulnerable to incompleteness because facts in the real world are countless and ever-evolving\cite{trivedi2017knowevolve}. In order to complete KGs, numerous methods referred as link prediction or knowledge graph completion have been designed to reduce the gap between KGs and real-world knowledge\cite{Guan2022N-ary}. In general, algorithms for link prediction can be divided into three groups: embedding-based method, logical-induction method and subgraph-based method. 

\begin{figure}[t]
\centering
\includegraphics[width=0.9\columnwidth]{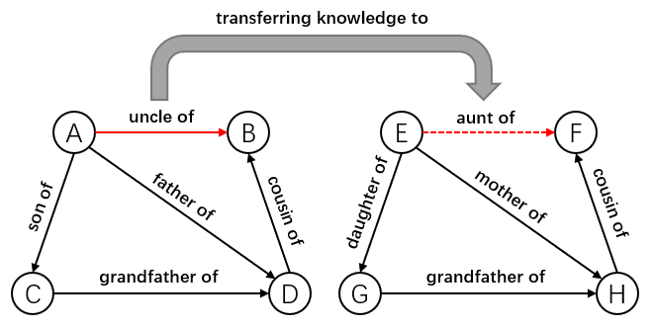} % Reduce the figure size so that it is slightly narrower than the column. Don't use precise values for figure width.This setup will avoid overfull boxes.
\caption{Illustration of predicting unseen relation using similar and familiar relation. The red solid arrow indicates the familiar target triplet in the training set. The red dotted arrow means unseen target triplet for prediction. Relations of these two triplets are similar in semantic meaning, resulting in similar extracted subgraphs.}.
\label{fig:intro}
\end{figure}

Although these groups of methods have shown promising performance, they all have some shortcomings caused by their techniques for processing KGs. Specifically, as embedding-based methods predict relation based on the learned embeddings of entities and relations, they are incapable of inferring relations between entities not presented in the training set, which are limited to transductive setting\cite{transductive}. Recently, researchers have extended embedding-based models to inductive versions\cite{Hamaguchi2017Knowledge,wang2019logic} which 
can make prediction on unseen entities without re-training. However, they require extra computation and are incapable of handling entities without additional information\cite{Xie2016Image,Xie2016RepresentationLO,Shi2017Open-World}. As for logical-induction methods\cite{Luis2015Fast,Yang2017Differentiable}, they induce the probabilistic logic rules by learning the statistical regularities and patterns hidden in the KGs\cite{Meilicke2018Fine-Grained,sadeghian2019drum}. Nevertheless, it is impractical to apply logical-induction models on large knowledge graph because of excessive computation on learning statistical regularities. Inspired by the fact that relation between two entities can be inferred according to local subgraph constructed by neighbor triplets, subgraph-based methods apply Graph Neural Network (GNN) on the extracted subgraph to obtain prediction. Despite the impressiveness of model performance, subgraph-based models cannot directly predict unseen relation as its embedding is unknown to the models, i.e., they are not completely inductive. Another drawback of existing subgraph-based models is that they ignore semantic information of relation and fail to attend to long-tail relations. In detail, as subgraph-based models independently assign learnable parameters for each relation, the parameters for a specific relation can be updated only when the inputted subgraph contains the relation. For long-tail relations, their parameters get updated for a few training steps, leading to weak expressive power of their embeddings and thus inaccurate prediction.

Aiming to address the above issues, we propose a novel subgraph-based inductive model named \textbf{Re}lation-dependent \textbf{Co}ntrastive \textbf{Le}arning (ReCoLe). More specifically, drawing from the recent advances in contrastive learning (CL), we devise a novel CL and an innovative sampling method for inductive relation prediction task. In order to further improve the role of relation, ReCoLe assigns special encoder for each relation and applies CL objective to pre-train the encoder. The CL objective enables ReCoLe to understand deep semantics of relations by minimizing the distance between the representations of subgraphs of similar target relations. In other words, after CL, a specific encoder has learned how to encode the extracted subgraph for the corresponding target relation into an effective graph-level representation. As for the novel sampling method, it is based on clustering algorithm which classifies all relations according to their semantics for selecting positive and negative samples for CL. The proposed cluster sampling equips ReCoLe with the ability to handle unseen relations. Generally, ReCoLe can understand an unseen relation by analogy with other familiar relations which has similar semantic meaning with the unseen relation. For instance, as shown in Fig~\ref{fig:intro}, two target relations $uncle\_of$ and $aunt\_of$ have similar semantic meaning. When the model handles the unseen relation $aunt\_of$, it can absorb learned knowledge from the familiar relation $uncle\_of$ to complete prediction. In contrast to previous efforts\cite{teru2020inductive,mai2021communicative} that fail to handle unseen relation, ReCoLe is completely inductive. Given an unseen target relation, ReCoLe can first sort it into the cluster which contains relation with similar semantics. Then it applies the corresponding encoder to process the unseen relation. Unlike previous subgraph-based methods which overlooks long-tail relations, our model ensures satisfactory performance on all relations and enhances the influence of relation as each relation owns an encoder pre-trained by contrastive learning.

In brief, our contributions are summarized as follows:
\begin{itemize}
    \item We introduce a new framework, ReCoLe, adapting contrastive learning with a novel cluster sampling method for inductive relation prediction. By pre-training encoder for each relation with CL, the role of relation gets improvement, which fits the nature of inductive setting.
    \item We propose a novel sampling method based on cluster algorithm, which not only makes our model completely inductive, but also improves the performance on long-tail relations. ReCoLe is capable of handling both unseen entities and relations.
    \item We evaluate ReCoLe on several commonly used inductive datasets. The experimental results suggest that our model significantly outperforms previous state-of-the-art inductive models in Hits@10 and AUC-PR metrics.
\end{itemize}

\section{Related Work}
\subsection{Transductive models} As representation learning on KGs are well developed, most previous models for knowledge graph completion focus on learning effective embeddings for entities and relations, which are called embedding based models\cite{WANG2021259,ZHANG2022109,NAYYERI2021530,ShenModeling2021}. Typical models including TransE\cite{TransE}, Dismult\cite{Distmult}, ComplEx\cite{ComplEx} and ConvE\cite{ConvE} apply simple computation or neural network to learn embedding for each triplet independently, failing to integrate abundant semantic and structural information of the graph into the embedding. In order to make up for this defect, recent embedding-based models\cite{SACN,R-GCN,nathani2019learning} utilize Graph Neural Network (GNN) to aggregate global information of the graph and produce more expressive embedding. However, all the above embedding-based models rely on the learned embedding to make prediction, leading to the limitation of transductive setting.

\subsection{Inductive models} Previous inductive models can be roughly divided into three groups: inductive embedding-based models, logical-induction models and subgraph-based models. The inductive embedding-based models insist the method of representation learning, trying to obtain embedding for unseen entities by external resources. For example, IKRL\cite{Xie2016Image} depends on the images of unseen entities to learn inductive embedding. Obviously, for entities without additional information, these inductive embedding-based models are impotent. To relieve this issue, PathCon\cite{PathCon} achieves embedding by a relational message passing which completely discards the semantic information of entities and attaches importance to relations. However, it is incapable of directly handling unseen relation as its embedding is unknown to the model.

As for logical-induction models, they focus on learning statistical regularities and patterns hidden in the KGs. For instance, RuleN\cite{Meilicke2018Fine-Grained} explicitly extracts path-based rules to induce probabilistic logic rules. Although they achieve impressive performance, they cannot be easily applied on large KGs due to excessive complexity of statistical regularities. Moreover, their rule-based nature causes the problem of less expressive power of the models.

Recently, subgraph-based models become influential advances in inductive relation prediction, e.g., GraIL\cite{teru2020inductive}, TACT\cite{chen2021topology}, CoMPILE\cite{mai2021communicative} and Meta-iKG\cite{zheng2021subgraphaware}, which infer relation based on the extracted subgraph. In general, their algorithms could be summarized as three steps: (i) extract subgraph surrounding the target triplet, (ii) initialize node (entity) embedding according to its relative position and parameterize edge (relation) embedding as learnable matrix which is updated by gradient descent, (iii) obtain the score of the target triplet using a module based on GNN. With semantic information of entities blocked, these models learn entity irrelevant rules and thus are inductive. Nevertheless, they can only predict relation that exists in the training set, i.e., they are not completely inductive for incapability of handling unseen relation. In addition, although these methods devise special module for modeling relation information, they have inferior performance in predicting long-tail relations which rarely present in the training set.

\subsection{Contrastive learning}
Contrastive learning is first proposed in computer vision tasks\cite{zhuang2019local, he2020momentum} which makes the transformed representations of different augmented images agree with each other. The key idea behind contrastive learning is that the generalization capability of the model could be improved after comparing augmentations created by different samples. As for the proposed relation-dependent contrastive learning, it aims to help our model further understand semantic for relations by comparing representations created by different relations. Most previous works relevant to contrastive learning in Natural Language Processing (NLP) focus on finding suitable agumentation for text\cite{giorgi2020declutr}, sentences\cite{wu2020clear} and KGs\cite{you2020graph}. It is worth noting that our model does not rigidly define novel augmentations, while devotes to selecting appropriate positive and negative samples for contrastive learning. In essence, two augmentations from the same sample are a positive pair and augmentations from different sample are a negative pair. We skip the definition of augmentations and directly design a novel method to select positive and negative samples.

\begin{figure*}[t]
\centering
\includegraphics[width=0.8\textwidth]{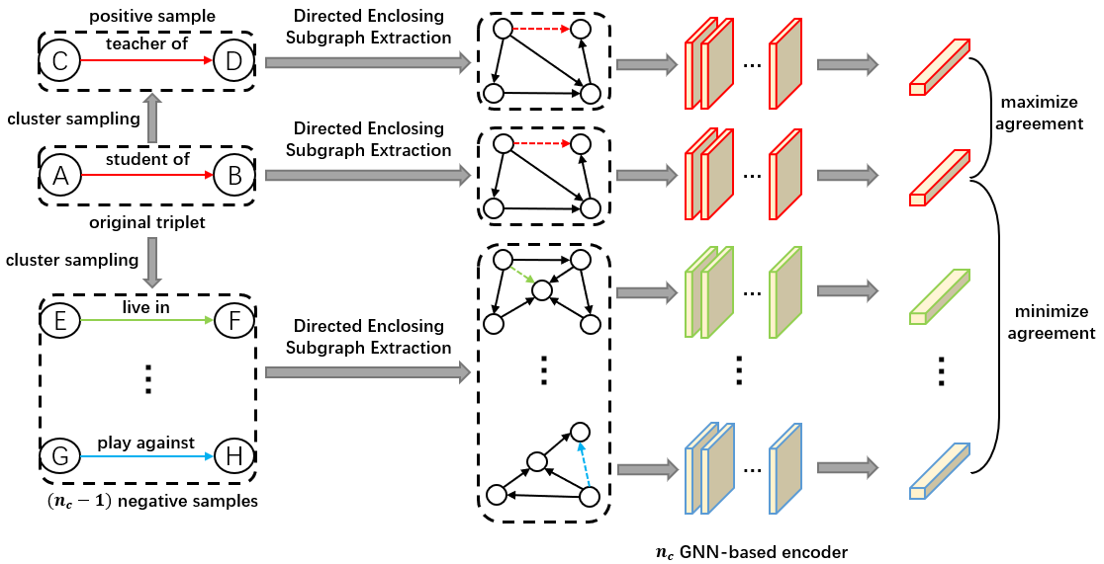}
\caption{The schematic diagram of our proposed ReCoLe. The target relation is marked with a color. Different colors indicate different clusters of relations. Note that encoders are also assigned with colors because a subgraph is processed by its corresponding encoder.}
\label{fig:alg}
\end{figure*}

\section{Algorithm}
The key idea behind our approach is to leverage contrastive learning to obtain pre-trained encoders which have learned how to encode the extracted subgraph surrounding a target triplet into graph-level representation. During contrastive learning, positive and negative samples are selected by our proposed cluster sampling which improves the ability of encoders and enables our model to predict unseen relation. We elaborate ReCoLe with its diagram illustrated in Fig~\ref{fig:alg}, which consists of four steps: (i) selecting positive and negative samples with cluster sampling, (ii) extracting the directed enclosing subgraphs of different target triplets, (iii) feeding the subgraphs into their corresponding GNN-based encoders, (iv) optimizing all the encoders with CL loss. After pre-training the encoders, we fine-tune ReCoLe with the inductive relation prediction task.

\subsection{Cluster Sampling}
For a target relation, we hope the model can find appropriate positive and negative samples, so that the encoder could learn statistical regularities hidden in the subgraph surrounding this target relation. In addition, we expect the model to pay enough attention to long-tail relations. To this end, we first classify all the relations according to their semantic meaning, then randomly choose triplets with relations in the same cluster and different clusters as positive sample and negative samples, respectively. Note that all the chosen triplets are from the same training set. For instance, in Fig.~\ref{fig:alg}, triplets $(A, student\_of, B)$ and $(C, teacher\_of, D)$ have similar semantic meanings of relations, resulting that they are clustered in the same group and the model regards them as a positive pair. To obtain semantic meaning of a relation, we first find Glove word embedding\cite{pennington2014glove} of each description word of the relation, then average all the embeddings as the semantic representation of this relation. We denote the Glove embedding of relations as $R_{glo}\in{\mathbb{R}^{n_r\times 300}}$ with $n_r$ being the number of relations. To reduce noise information and facilitate clustering, we use t-SNE\cite{t-SNE} model to obtain relation embeddings which have lower and the same dimensionality:
\begin{equation} \label{t-sne}
R = {\rm t\mbox{-}SNE}(R_{glo})
\end{equation}
where $R\in{\mathbb{R}^{n_r\times d_r}}$ means the sets of relation embeddings after dimensionality reduction with $d_r$ indicating the reduced dimensionality. As for the clustering algorithm, we choose K-Means to obtain $n_c$ groups of relations:
\begin{equation} \label{K-Means}
R_1, R_2, ..., R_{n_c} = {\rm K\mbox{-}Means}(R, n_c)
\end{equation}
where $R_1, R_2, ..., R_{n_c}$ are different groups of relations. For selecting a positive sample of the original triplet, we first randomly choose a relation in the same cluster, then randomly select a triplet with this chosen relation from the training set. With the above similar operation, we can obtain $n_c-1$ negative samples from the other $n_c-1$ clusters. By this proposed cluster sampling, ReCoLe divides all the relations into different groups according to semantics and distributes different encoders to them. As a result, one encoder will learn specific knowledge about its group of relations after contrastive learning, which enhances the influence of relations. In addition, as relations in the same group share the same encoder, ReCoLe could better understand long-tail relations with complementary knowledge of relations in the same group. Moreover, the cluster sampling equips our model with the ability of predicting unseen relation. Specifically, once the unseen relation is classified into an appropriate cluster, the model can apply the corresponding pre-trained encoder to produce graph-level representation and get prediction.

\subsection{Directed Enclosing Subgraph Extraction}
In this section, we first introduce how to obtain a directed enclosing subgraph, then define the embeddings of nodes and edges in the subgraph. The subgraph extraction method aims to extract a subgraph surrounding a target triplet from the original KG. Specifically, given a target triplet $(h, r, t)$, ReCoLe first finds all the $m$-hop outgoing neighbours of $h$ and incoming neighbours of $t$. Note that $h$ is the 1-hop incoming neighbour of $t$ and correspondingly, $t$ is the 1-hop outgoing neighbour of $h$. Then, the model computes the intersection between the set of outgoing neighbours and the set of incoming neighbours, i.e., common nodes between the outgoing neighbours and incoming neighbours. Finally, the subgraph is constructed by these common nodes and the corresponding edges between these nodes. The target edge (relation) will be dropped in order to prevent labels from leaking. In the extracted subgraph, all the neighbours of $h$ and $t$ are outgoing and incoming, respectively. In other words, $h$ is always the head entity in the triplets where it is located. Similarly, $t$ is always the tail entity. As a result, the extracted subgraph is enclosing and has a direction, which we call directed enclosing subgraph. 

As the model is required to be entity-independent, we define node embedding based on its relative position in the subgraph, which contains no semantic information of the node and represents structural information of the subgraph. The relative position of node $i$ can be represented by the minimum distances from the node $i$ to the head $h$ and tail $t$ nodes of target triplet in the subgraph:
\begin{equation} \label{node embedding}
N_i = {\rm one\mbox{-}hot}(d_{hi}) \oplus {\rm one\mbox{-}hot}(d_{it})
\end{equation}
where $N_i\in{\mathbb{R}^{2(m+2)}}$ denotes the embedding of node $i$ with $m$ is the number of hops defined above. $d_{hi}$ and $d_{it}$ represent the minimum distances from node $i$ to the target head $h$ and tail $t$, respectively. For edge embedding $E_j$ representing the $j$th relation type, it is initialized as an one-hot vector, which is consistent with the definition of node embedding to facilitate the message passing mechanism in the subsequent encoder. Therefore, given a directed enclosing subgraph, we can obtain the set of node embedding $N\in{\mathbb{R}^{n_o \times 2(m+2)}}$ and the set of edge embedding $E\in{\mathbb{R}^{n_e \times n_r}}$ with $n_o$ and $n_e$ denoting the numbers of nodes and edges in the subgraph and $n_r$ being the total number of relation types in the dataset. Moreover, the connections in the subgraph are reflected by head-to-edge and tail-to-edge adjacency matrices $A^{he}\in\mathbb{R}^{n_o\times n_e}$ and $A^{te}\in\mathbb{R}^{n_o\times n_e}$ which map the head and tail nodes to the corresponding edge, respectively.

\subsection{GNN-Based Encoder}
Finishing extracting subgraph and defining embedding, we need to project both the node and edge embeddings to the same dimension:
\begin{equation} \label{initialization_node}
N^0 = NW^n
\end{equation}
\begin{equation} \label{initialization_edge}
E^0 = EW^e
\end{equation}
where $N^0\in\mathbb{R}^{n_o\times d}$ and $E^0\in\mathbb{R}^{n_e\times d}$ denote the initial node and edge embeddings, i.e., the embeddings in the 0th layer in the following Graph Neural Network (GNN), respectively. In general, we can apply various kinds of GNN to encode the subgraph into graph-level representation. Aiming to improve the role of relation, we choose the relational message passing mechanism\cite{PathCon} as the main module of the encoder. Specifically, in each iteration $k$, the edge embedding first absorbs information from the head and tail nodes connected at both ends. Then, the node embedding is updated based on the edges connected to the node:
\begin{equation} \label{edge update}
E^{k+1} = E^k + (A^{he})^TN^k + (A^{te})^TN^k
\end{equation}
\begin{equation} \label{node update}
N^{k+1} = f((A^{he}E^{k+1}+A^{te}E^{k+1}+N^k)W^k)+b^k)
\end{equation}
where $f$ represents the non-linear activation functions. To emphasize the structural information of the subgraph, we add initial node embedding and concatenate the target head and tail node embeddings as the final representation of the subgraph:
\begin{equation} \label{}
V = N^0 + N^K
\end{equation}
\begin{equation} \label{}
H = {\rm Concat}(v_h, v_t)
\end{equation}
where K is the total iterations of the relational message passing. $v_h$ and $v_t$ denote the target head and tail node embeddings obtained from $V$, respectively, with $H\in\mathbb{R}^{2\times d}$ being the final representation of the subgraph. Note that an encoder of a specific relation cluster has its own learnable parameters and target relations with similar semantic meaning share the same encoder. As a result, each encoder could effectively learn specific knowledge of its relation cluster, which improves the influence of relation on the prediction.

\subsection{Contrastive Loss}
As we obtain the representations of positive and negative samples, we can apply the contrastive loss function to train the encoders. The loss for a target triplet is defined as:
\begin{equation} \label{}
\ell = -{\rm log}\frac{{\rm exp(sim}(H_{tar},H_{pos})/\tau}{\sum_j^{n_c-1}{\rm exp(sim}(H_{tar}, H_{neg}^j)/\tau)}
\end{equation}
where ${\rm sim}(H_1, H_2)=H_1^TH_2^T/\Arrowvert H_1\Arrowvert \Arrowvert H_2\Arrowvert$ is the cosine similarity function with $\tau$ being the temperature parameter. Finishing pre-training all the encoders with contrastive loss, we apply soft margin loss to fine-tune the model. Specifically, given a target triplet $(h,r,t)$, the model first extracts its subgraph, then produces representation by an encoder which depends on the relation $r$. Finally, the representation is fed into a fully connected layer to obtain the final prediction. Thus, the model is fine-tuned by comparing the prediction and the ground-truth label with soft margin loss.

\section{Experiments}
In this section, we first present the details about datasets, baseline models and experimental settings, then compare the performance of ReCoLe with the baselines. Finally, we conduct some additional experiments and provide vivid illustrations to prove the superiority of the proposed relation-dependent contrastive learning and cluster sampling.

\begin{table*}[t]
\centering
 \caption{\textbf{Comparison with baselines on two inductive datasets.} The bold means the best performance. Note that each dataset owns four versions denoted as v1, v2, v3 and v4.}
 \label{tab:performance}
\resizebox{0.8\textwidth}{!}
{\begin{tabular}{c|cccc|cccc}
 \hline
 \multirow{2}{*}{Models} & \multicolumn{4}{c|}{FB15k-237} & \multicolumn{4}{c}{NELL-995}\cr
  & v1 & v2 & v3 & v4 & v1 & v2 & v3 & v4\cr
 \hline
 \multicolumn{9}{c}{AUC-PR}\cr
 \hline
 Relational GAT & 81.57 & 84.03 & 81.48 & 81.62 & 68.95 & 80.78 & 81.89 & 81.64\cr
 RuleN & 79.60 & 82.67 & 83.03 & 84.01 & 67.12 & 80.52 & 73.91 & 77.07\cr
 PathCon & 72.92 & 83.53 & 80.00 & 80.38 & 67.82 & 79.86 & 76.62 & 74.50 \cr
 GraIL & 80.45 & 83.66 & 84.35 & 83.08 & 69.35 & 85.04 & 84.43 & 80.19\cr
 CoMPILE & 79.95 & 83.56 & 83.97 & 83.87 & 68.36 & 85.50 & 84.04 & 79.89\cr
 Meta-iKG (MAML) & 80.31 & 82.95 & 82.52 & 84.23 & 72.12 & 84.11 & 82.47 & 79.25\cr
 Meta-iKG (Meta-SGD) & 81.10 & 84.26 & 84.57 & 83.70 & 72.50 & 85.97 & 84.05 & 81.24\cr
 ReCoLe & \textbf{82.77} & \textbf{85.80} & \textbf{86.58} & \textbf{86.30} & \textbf{72.82} & \textbf{86.66} & \textbf{86.40} & \textbf{85.86}\cr
 \hline
 \multicolumn{9}{c}{Hits@10}\cr
 \hline
 Relational GAT & 64.78 & 71.58 & 68.29 & 67.64 & 58.64 & 73.72 & 75.43 & 74.11\cr
 RuleN & 65.35 & 71.68 & 67.84 & 70.53 & 53.70 & 69.77 & 64.29 & 57.92\cr
 PathCon & 55.65 & 73.74 & 63.16 & 65.23 & 54.32 & 70.93 & 61.95 & 61.29 \cr
 GraIL & 66.52 & 73.82 & 70.15 & 68.30 & 55.56 & 76.40 & 75.66 & 71.24\cr
 CoMPILE & 66.52 & 72.37 & 69.77 & 70.27 & 62.35 & 76.51 & 75.58 & 68.19\cr
 Meta-iKG (MAML) & 66.52 & 72.37 & 68.81 & 74.32 & 60.49 & 74.07 & 77.99 & 71.63\cr
 Meta-iKG (Meta-SGD) & 66.96 & 74.08 & 71.89 & 72.28 & 64.20 & 77.91 & 77.41 & 73.12\cr
 ReCoLe & \textbf{71.30} & \textbf{76.58} & \textbf{76.06} & \textbf{75.46} & \textbf{64.81} & \textbf{83.60} & \textbf{79.08} & \textbf{79.94}\cr
 \hline
 \end{tabular}}
 \vspace{-0.3cm}
\end{table*}

\subsection{Datasets}
FB15k-237\cite{FB15k-237} and NELL-995\cite{nell} are popular datasets for knowledge graph completion. In GraIL, the authors create inductive versions for these two datasets by sampling disjoint subgraphs from the original KGs. Specifically, they sample two disjoint subgraphs for each dataset: \textit{train-graph} and \textit{test-graph}. These two subgraphs have the same set of relations but have a disjoint set of entities. For robust evaluation, the authors create four versions for each dataset by sampling different \textit{train-graph} and \textit{test-graph}. Recently, some studies\cite{mai2021communicative} introduce refined inductive versions for these two datasets which exclude empty-subgraph triplets to achieve more scientific and reasonable evaluation. Empty-subgraph triplet means that we cannot find any valid edge in the extracted enclosing subgraph of the triplet. Therefore, for subgraph-based models, the prediction on empty-subgraph triplet is meaningless. Moreover, as we construct the negative triplet by replacing the original head or tail nodes, there will be much more empty-subgraph negative triplets in the original dataset. To this end, and to be consistent with prior works, we evaluate our model on the refined inductive dataset.

\subsection{Baseline Models}
We compare ReCoLe with six baseline models including Relational GAT\cite{graph_attention}, RuleN\cite{Meilicke2018Fine-Grained}, PathCon\cite{PathCon}, GraIL\cite{teru2020inductive}, CoMPILE\cite{mai2021communicative} and Meta-iKG\cite{zheng2021subgraphaware}. Relational GAT is an adapted model which extends Graph Attention Network (GAT) to perform inductive relation reasoning. Before updating node embedding for the next iteration, GAT will assign an adaptive attention weight for each edge. Therefore, more informative edge will be highlighted during message passing. RuleN is an end-to-end differentiable method aming to learn logical rules from KGs. However, it pays no attention to the neighbors surrounding the target triplet, leading to inaccurate prediction on sparse KG. PathCon adapts a relational message passing mechanism to learn two kinds of neighborhood topology called relational context and relational paths. GraIL, CoMPILE and Meta-iKG are subgraph-based models which first extract subgraph surrounding a target triplet, then make prediction based on the extracted subgraph. CoMPILE focuses on improving the message passing mechanism to obtain effective interactions between edges and nodes. In addition, it proposes to extract directed enclosing subgraphs to solve the problem of asymmetric/anti-symmetric relations. As previously stated, they are all incapable of handling unseen relation and pay insufficient attention to the role of relation.

\subsection{Experimental Settings}\label{sec:exper_detail}
We develop our model on Pytorch and conduct experiment on a Nvidia GTX 1080 Ti GPU. To determine the hyperparameters, we perform a basic grid search to find the best hyperparameter setting. In accordance with prior works, the reported metrics are AUC-PR and Hits@10. The former is computed by comparing the scores of test triplet and the corresponding sampled negative triplet. Larger AUC-PR indicates better model performance because more test triplets are scored higher than negative triplets. To compute Hits@10, we first create 50 negative triplets for each test triplet by replacing the head or tail with other entities. Then the test triplet and these sampled negative triplets are fed into the models to obtain their scores. Hits@10 reflects the proportion of the test triplets which rank in top 10 according to their scores.

\subsection{Results and Disccussions}
\subsubsection{Comparison with Baselines}
Table~\ref{tab:performance} shows the overall performance achieved by ReCoLe and the baselines, which indicates consistent improvement on the majority of the inductive datasets in terms of both the AUC-PR and Hits@10 metrics. It is worth noting that our model outperforms the previous state-of-the-art subgraph-based model Meta-iKG by a considerable margin. Specifically, for the Hits@10 metric on four versions of FB15k-237 dataset, the absolute gain of ReCoLe against Meta-iKG (MAML) model are 4.78\%, 4.21\%, 7.25\% and 1.14\%, respectively. Comparing with Meta-iKG(Meta-SGD) on the basis of AUC-PR, our model achieves an average improvement of 1.96\%. As information of relation is vital for inductive relation prediction, we attribute this improvement to our proposed relation-dependent contrastive learning which can improve the role of relation. In addition, there are moderate number of triplets containing all the long-tail relations in the dataset. So, more accurate prediction on long-tail relations is an important factor in boosting model performance. In contrast to CoMPILE which overlooks this factor, ReCoLe adapts contrastive learning to pre-train the GNN-based encoders with the help of the proposed cluster sampling. As a result, the encoders can produce effective representations for triplets with long-tail relation. From Table~\ref{tab:performance} we find that ReCoLe outperforms CoMPILE by 2.46\%, 7.09\%, 3.5\% and 11.75\% on Hits@10 metric on NELL-995, respectively, proving the above statement about the benefit brought by sufficient attention to long-tail relations. Comparing with logical-induction model RuleN, ReCoLe also achieves significant improvement across almost all the metrics, indicating subgraph-based method is a more practicable and effective choice for inductive relation prediction. As previously stated, logical-induction model is incapable of handling large KGs with excessive complexity of statistical regularities. In contrast, ReCoLe can be easily applied on large KGs as it is a subgraph-based method. More importantly, it achieves more satisfactory performance than logical-induction model, e.g., an average gain of 15.44\% on Hits@10 on NELL-995.

\begin{figure*}[t]
\centering
\includegraphics[width=0.8\textwidth]{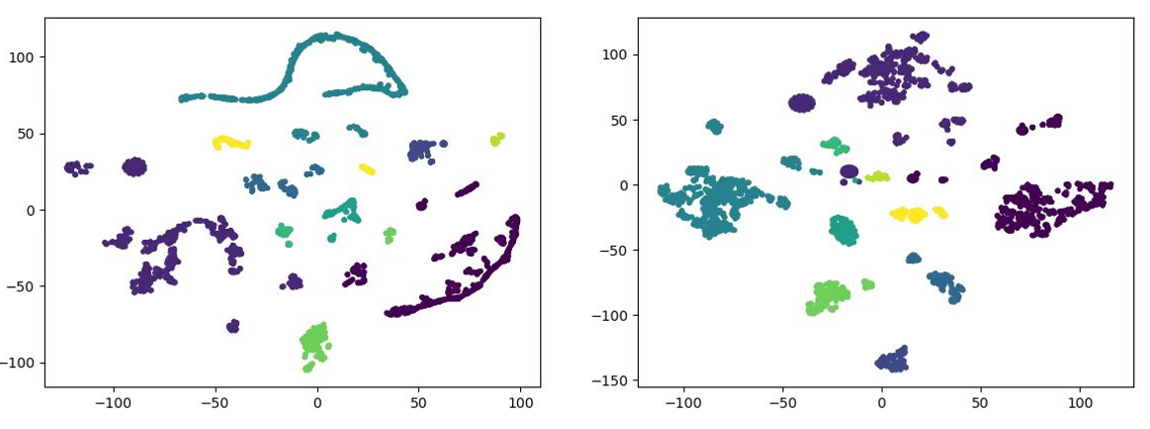} % Reduce the figure size so that it is slightly narrower than the column.
\caption{Visualization of features produced by untrained (left) and pre-trained (right) encoders on FB15k-237-v1 dataset. Each point represents a representation which is processed by t-SNE model. We distinguish different groups of relations with different colors.}
\label{fig:vis}
\end{figure*}

\begin{table}[t]
\centering
  \caption{An ablation study on the benefit of contrastive learning and cluster sampling using FB15k-237.}  
\label{tab:ablation}  
\resizebox{0.48\textwidth}{!}
    {\begin{tabular}{c|cc|cc}  
    \hline
    \multirow{2}{*}{Models} & \multicolumn{2}{c|}{v1} & \multicolumn{2}{c}{v2} \cr
    & AUC-PR & Hits@10 & AUC-PR & Hits@10 \cr
    \hline
    %GRE & 80.53 & 65.22 & 84.43 & 72.24\cr  
    W/O Cluster Sampling  & 75.44 & 56.96 & 80.76 & 70.13\cr
    W/O Contrastive Learning & 81.21 & 68.26 & 84.22 & 74.61\cr
    W/O Positive Sample & 81.99 & 65.65 & 81.20 & 67.63 \cr
    W/O Negative Sample & 73.97 & 56.09 & 79.72 & 69.61\cr
    \hline
    ReCoLe & \textbf{82.77} & \textbf{71.30} & \textbf{85.80} & \textbf{76.58} \cr
    \hline
    \end{tabular}}
\end{table}

\subsubsection{Ablation Study}
In order to verify the effectiveness of our proposed contrastive learning and cluster sampling, we conduct ablation study on FB15k-237 datasets as Table~\ref{tab:ablation} shows. For the case of W/O Cluster Sampling, we randomly classify relations into different clusters. As for W/O Contrastive Learning, we first initialize encoders with learnable parameters, then directly train the model with relation prediction task. The cases of W/O Positive Sample and W/O Negative Sample are designed by discarding positive and negative samples of cluster sampling, respectively. Compared to the model W/O Cluster Sampling, ReCoLe achieves 7.33\% and 4.90\% improvement on AUC-PR metric in FB15k-237-v1 and FB15k-237-v2 datasets. The reason behind this improvement is that ReCoLe can transfer knowledge from familiar relations to long-tail relations with similar semantic meanings, as similar relations are classified into the same cluster. The prediction for long-tail relation becomes more exact and reliable. In addition, the improvement proves that our proposed cluster sampling can select appropriate positive and negative samples for contrastive learning, facilitating the following encoders to learn mutual knowledge carried in subgraphs comprised of similar relations. Without contrastive learning (see the case of W/O Contrastive Learning), we find significant decline of the model performance, which indicates the pre-training by our proposed relation-dependent contrastive learning is an effective means to help our model learn prior knowledge about relation. The absorbed prior information strengthens the role of relation in making prediction , and thus boosts model performance. From the case of W/O Positive Sample and W/O Negative Sample, we can observe that the absence of either positive or negative samples will lead to worse performance. Comparing with other devised situations for ablation study, the absence of negative samples causes the most significant decline. It may be because the negative samples are the most important components in our proposed contrastive learning, as they are responsible for training most of the encoders.

\subsubsection{Visualization for Contrastive Learning}
To further illustrate the effectiveness of the proposed relation-dependent contrastive learning, we visualize and compare the output features from encoders with and without contrastive learning. Specifically, we feed all the triplets in FB15k-237-v1 datasets into ReCoLe which is pre-trained by contrastive learning. All the output representations are processed by a t-SNE\cite{t-SNE} model to obtain a 2D plot. Similarly, we also apply untrained ReCoLe and t-SNE model to produce representations for comparison. The visualized results are shown in Fig~\ref{fig:vis} where the left and right plots are for representations from untrained and pre-trained ReCoLe, respectively. Different colors represent different groups of relations. For example, the representations of target triplets $(A, student\_of, B)$ and $(C, classmate\_of, D)$ are assigned the same color as their relations are similar.

From the left plot in Fig.~\ref{fig:vis}, we find that most points with the same color are distributed as lines, e.g. the points with cyan color in the top of the plot. In contrast, the right plot shows apparent clusters with different colors, indicating target triplets with similar relations have similar representations. Specifically, as mentioned earlier, the extracted subgraphs of target triplets with similar relations have similar structures. Therefore, their representations have correlation which the model should be aware of. With the proposed relation-dependent contrastive learning, ReCoLe can recognizes this correlation and produces effective representations for similar triplets. As the contrastive learning aims to assist ReCoLe to learn underlying message of relations, the role of relation for the final prediction is significantly improved, which is a key to further boost the model performance. Moreover, when ReCoLe handle triplet with long-tail relation, i.e, only a few samples contain this relation, the model can produce superior representation for it. Because the long-tail relation is aided by other relations with similar semantic meaning.

\subsubsection{Stronger Performance on Long-tail Relations}
As previously defined, long-tail relations are the relations which rarely appear in the training set. Previous subgraph-based models assign separate one-hot or parameterized learnable embedding for each relation, leading to inadequate training and thus weak expressive power for embedding of long-tail relation. Instead of directly learning embedding for long-tail relations, ReCoLe adapts contrastive learning to transfer complementary knowledge from well-studied relations to long-tail relations. In order to further prove the superior capability of ReCoLe to handle long-tail relations, we conduct additional experiments with the following steps: (i) count the number of occurrences for each relation in the training set, (ii) extract the relation whose number of occurrences is less than a predefined threshold, (iii) evaluate our pre-trained model and a baseline model, PathCon, on the extracted long-tail relations and compare the performance on AUC metric. 

\begin{table}[t]
\centering
  \caption{AUC Performance on long-tail relations with different thresholds on FB15k-237. Note that the threshold represents the maximum number of occurrences of the extracted long-tail relations in the dataset.}  
\label{tab:long-tail relation}  
\resizebox{0.48\textwidth}{!}
    {\begin{tabular}{c|c|cccc}  
    \hline
    \multirow{2}{*}{Dataset} & \multirow{2}{*}{Models}
    & \multicolumn{4}{c}{Threshold} \cr
     & & 20 & 30 & 40 & 50\cr
    \hline
    \multirow{2}{*}{FB15k-237-v1} & PathCon & 81.65 & 80.20 & 78.61 & 79.48 \cr
     & ReCoLe & 83.25 & 83.55 & 79.88 & 80.94 \cr
    \hline
    \multirow{2}{*}{FB15k-237-v2} &  PathCon & 80.52 & 81.20 & 80.68 & 82.14\cr
     & ReCoLe & 90.37 & 91.25 & 89.65 & 90.86\cr
    \hline
    \end{tabular}}
\end{table}

\begin{table}[t]
\centering
  \caption{The total number of triplets containing long-tail relations with different thresholds. As the threshold increases, the number of triplets becomes larger because more long-tail relations are taken into account.}  
\label{tab:sample num}  
\resizebox{0.35\textwidth}{!}
    {\begin{tabular}{c|cccc}  
    \hline
    \multirow{2}{*}{Dataset} & \multicolumn{4}{c}{Threshold} \cr
     & 20 & 30 & 40 & 50\cr
    \hline
     FB15k-237-v1 & 107 & 137 & 192 & 237 \cr
    \hline
     FB15k-237-v2 & 120 & 152 & 196 & 276\cr
    \hline
    \end{tabular}}
\end{table}

Table~\ref{tab:long-tail relation} and Table~\ref{tab:sample num} indicate the performance on long-tail relations and the total number of triplets containing these long-tail relations with different thresholds, respectively. From Table~\ref{tab:long-tail relation} we can infer that ReCoLe is more competent than PathCon in handling long-tail relations. Specifically, for FB15k-237-v2 dataset, ReCoLe achieves nearly 10\% improvement for all the thresholds comparing with PathCon. Although some relations have only appeared for a few times during the training process, our model can make accurate predictions on them, e.g., 90.37 and 91.25 AUC for thresholds 20 and 30 on FB15k-237-v2 dataset. Moreover, with the increase of threshold, our model keeps satisfactory and stable performance. For instance, ReCoLe achieves an average AUC performance of 90.53 for all the thresholds on FB15k-237-v2 dataset and the gap between the maximum and the minimum values is 1.6. This implies that ReCoLe has a deep understanding of long-tail relations after contrastive learning. More specifically, by learning the underlying shared knowledge of triplets containing relations with similar semantics and distinguishing subgraphs composed of triplets containing relations with different semantics, all the GNN-based encoders in ReCoLe can produce effective representations when handling long-tail relations.

\begin{figure*}[t]
\centering
\includegraphics[width=0.8\textwidth]{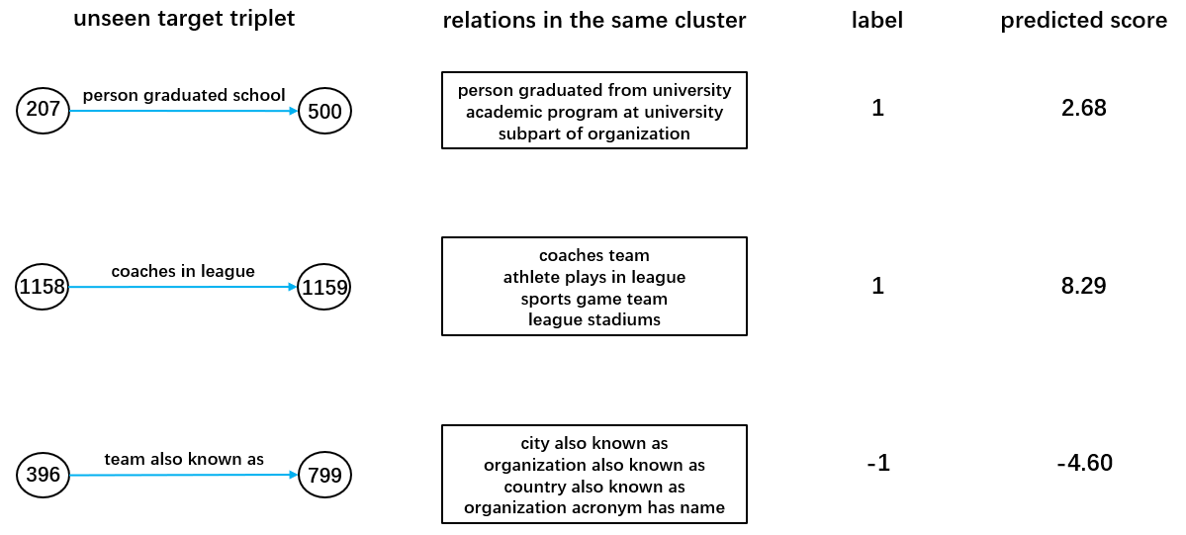} % Reduce the figure size so that it is slightly narrower than the column.
\caption{Visualization of the performance on unseen relation. The label indicates whether the fact represented by the triplet is true. Higher predicted score means that the model regards the triplet as true in a higher degree. We also present other relations in the same cluster as the unseen target relation, which provides significant information for the prediction. Note that all the nodes (entities) are represented by their ids as the experiment is conducted in inductive setting.}
\label{fig:vis_unseen}
\end{figure*}

\subsubsection{Case Study for Unseen Relations}
Owing to the adapted relation-dependent contrastive learning and cluster sampling, ReCoLe is completely inductive, i.e., it is capable of making prediction for unseen relation which is not included in the training set. Specifically, there are multiple GNN-based encoders in our model. Each encoder accounts for processing a group of relations which have similar semantic meaning. With proper positive and negative samples selected by cluster sampling and contrastive learning, all the encoders can learn common knowledge of relations in the same group and produce effective representations for their subgraphs. Therefore, given an unseen target relation, we can find an appropriate encoder for it according to its semantic meaning and obtain the prediction. To further validate the capability of ReCoLe to handle unseen relation, we conduct an additional experiment as follows: (i) eliminate all the triplets with a specific relation in the training set and train our model, (ii) feed the eliminated triplets into the pre-trained model and compare the ground-truth label and the prediction. The experimental results are visualized in Fig~\ref{fig:vis_unseen}.

From Fig~\ref{fig:vis_unseen} we can observe that our model makes accurate predictions for all the cases including positive and negative samples, verifying the effectiveness of the proposed method to handle unseen relation. To be specific, the model successfully matches an appropriate encoder for the unseen target triplet. For instance, it feeds the triplet with relation `team also known as' to the encoder which has processed relations `city/organization/country also known as' in the training stage. Due to the fact that similar target relations own similar subgraphs, the model can refer to prior knowledge provided by these similar relations to handle the unseen target relation. In addition, from the column of `relation in the same cluster', we can infer that the proposed cluster sampling correctly sorts all the relations into different clusters, which makes significant contribution to learning deep semantic of relations. For example, the relations `coaches team', `athlete plays in league', `sports game team' and `league stadiums' are all phrases about sports. Our model is aware of this fact and puts them in the same cluster. Moreover, it is worth noting that the absolute predicted scores for `coaches in league' ($8.29$) and `team also known as' ($4.60$) are higher than that of `person graduated school' ($2.68$). We surmise that the difference of the scores is caused by the similarity of relations in the same cluster. In other words, the more similar the semantics of an unknown relation to those of other relations within the cluster, the more confident the model is in predicting. For instance, besides the relation `person graduated from university', `academic program at university' and `subpart of organization' are also regarded as similar relations for `person graduated school'. Obviously, the semantic similarity between these two relations and the target relation is not as large as that between `person graduated from university' and the target relation. While in the cases of `coaches in league' and `team also known as', all the relations in the same cluster are highly consistent in semantics. Therefore, higher semantic similarity among the relations in the same cluster will result in better performance on unseen relations.

\section{Conclusion}
In this paper, we develop a model called ReCoLe for inductive relation prediction where relation information is crucial for prediction. To strengthen the role of relation, we adapt a popular frame work, contrastive learning, to pre-train encoders for different relations. As contrastive learning aims to minimize the agreement between representations of similar triplets, the encoder can effectively learn specific knowledge of the set of similar relations. To equip the model with the ability of handling unseen relations, we propose a novel sampling method based on cluster sampling algorithm which cooperates with contrastive learning to make ReCoLe completely inductive. Experimental results on frequently used datasets show that ReCoLe achieves SOTA performance on inductive relation prediction task. The additional experiments prove the superiority of ReCoLe to handle long-tail and unseen relations.

%\section*{Acknowledgments}
%This should be a simple paragraph before the References to thank those individuals and institutions who have supported your work on this article.

%\begin{thebibliography}{1}
\bibliographystyle{IEEEtran}
\bibliography{./KG.bib}

\begin{thebibliography}{10}
\providecommand{\url}[1]{#1}
\csname url@samestyle\endcsname
\providecommand{\newblock}{\relax}
\providecommand{\bibinfo}[2]{#2}
\providecommand{\BIBentrySTDinterwordspacing}{\spaceskip=0pt\relax}
\providecommand{\BIBentryALTinterwordstretchfactor}{4}
\providecommand{\BIBentryALTinterwordspacing}{\spaceskip=\fontdimen2\font plus
\BIBentryALTinterwordstretchfactor\fontdimen3\font minus
  \fontdimen4\font\relax}
\providecommand{\BIBforeignlanguage}[2]{{%
\expandafter\ifx\csname l@#1\endcsname\relax
\typeout{** WARNING: IEEEtran.bst: No hyphenation pattern has been}%
\typeout{** loaded for the language `#1'. Using the pattern for}%
\typeout{** the default language instead.}%
\else
\language=\csname l@#1\endcsname
\fi
#2}}
\providecommand{\BIBdecl}{\relax}
\BIBdecl

\bibitem{recommendation}
F.~Zhang, N.~J. Yuan, D.~Lian, X.~Xie, and W.-Y. Ma, ``Collaborative knowledge
  base embedding for recommender systems,'' in \emph{Proceedings of the 22nd
  ACM SIGKDD international conference on knowledge discovery and data mining},
  2016, pp. 353--362.

\bibitem{semantic_research}
C.~Xiong, R.~Power, and J.~Callan, ``Explicit semantic ranking for academic
  search via knowledge graph embedding,'' in \emph{Proceedings of the 26th
  international conference on world wide web}, 2017, pp. 1271--1279.

\bibitem{question_answering}
W.~Cui, Y.~Xiao, H.~Wang, Y.~Song, S.-w. Hwang, and W.~Wang, ``Kbqa: Learning
  question answering over qa corpora and knowledge bases,'' \emph{Proc. VLDB
  Endow.}, vol.~10, no.~5, p. 565–576, jan 2017.

\bibitem{trivedi2017knowevolve}
R.~Trivedi, H.~Dai, Y.~Wang, and L.~Song, ``Know-evolve: Deep temporal
  reasoning for dynamic knowledge graphs,'' in \emph{international conference
  on machine learning}.\hskip 1em plus 0.5em minus 0.4em\relax PMLR, 2017, pp.
  3462--3471.

\bibitem{Guan2022N-ary}
S.~Guan, X.~Jin, J.~Guo, Y.~n. Wang, and X.~Cheng, ``Link prediction on n-ary
  relational data based on relatedness evaluation,'' \emph{IEEE Transactions on
  Knowledge and Data Engineering}, pp. 1--1, 2021.

\bibitem{transductive}
Z.~Yang, W.~Cohen, and R.~Salakhudinov, ``Revisiting semi-supervised learning
  with graph embeddings,'' in \emph{International conference on machine
  learning}.\hskip 1em plus 0.5em minus 0.4em\relax PMLR, 2016, pp. 40--48.

\bibitem{Hamaguchi2017Knowledge}
T.~Hamaguchi, H.~Oiwa, M.~Shimbo, and Y.~Matsumoto, ``Knowledge transfer for
  out-of-knowledge-base entities: A graph neural network approach,'' in
  \emph{Proceedings of the 26th International Joint Conference on Artificial
  Intelligence}, 2017, p. 1802–1808.

\bibitem{wang2019logic}
P.~Wang, J.~Han, C.~Li, and R.~Pan, ``Logic attention based neighborhood
  aggregation for inductive knowledge graph embedding,'' in \emph{Proceedings
  of the AAAI Conference on Artificial Intelligence}, vol.~33, no.~01, 2019,
  pp. 7152--7159.

\bibitem{Xie2016Image}
R.~Xie, Z.~Liu, H.~Luan, and M.~Sun, ``Image-embodied knowledge representation
  learning,'' in \emph{Proceedings of the 26th International Joint Conference
  on Artificial Intelligence}, ser. IJCAI'17.\hskip 1em plus 0.5em minus
  0.4em\relax AAAI Press, 2017, p. 3140–3146.

\bibitem{Xie2016RepresentationLO}
R.~Xie, Z.~Liu, J.~Jia, H.~Luan, and M.~Sun, ``Representation learning of
  knowledge graphs with entity descriptions,'' in \emph{Proceedings of the AAAI
  Conference on Artificial Intelligence}, vol.~30, no.~1, 2016.

\bibitem{Shi2017Open-World}
B.~Shi and T.~Weninger, ``Open-world knowledge graph completion,'' in
  \emph{Thirty-Second AAAI Conference on Artificial Intelligence}, 2018.

\bibitem{Luis2015Fast}
L.~Gal{\'a}rraga, C.~Teflioudi, K.~Hose, and F.~M. Suchanek, ``Fast rule mining
  in ontological knowledge bases with amie +,'' \emph{The VLDB Journal},
  vol.~24, no.~6, pp. 707--730, 2015.

\bibitem{Yang2017Differentiable}
F.~Yang, Z.~Yang, and W.~W. Cohen, ``Differentiable learning of logical rules
  for knowledge base reasoning,'' in \emph{Proceedings of the 31st
  International Conference on Neural Information Processing Systems}, 2017, p.
  2316–2325.

\bibitem{Meilicke2018Fine-Grained}
C.~Meilicke, M.~Fink, Y.~Wang, D.~Ruffinelli, R.~Gemulla, and
  H.~Stuckenschmidt, ``Fine-grained evaluation of rule-and embedding-based
  systems for knowledge graph completion,'' in \emph{International semantic web
  conference}.\hskip 1em plus 0.5em minus 0.4em\relax Springer, 2018, pp.
  3--20.

\bibitem{sadeghian2019drum}
A.~Sadeghian, M.~Armandpour, P.~Ding, and D.~Z. Wang, ``Drum: End-to-end
  differentiable rule mining on knowledge graphs,'' 2019.

\bibitem{teru2020inductive}
K.~Teru, E.~Denis, and W.~Hamilton, ``Inductive relation prediction by subgraph
  reasoning,'' in \emph{International Conference on Machine Learning}.\hskip
  1em plus 0.5em minus 0.4em\relax PMLR, 2020, pp. 9448--9457.

\bibitem{mai2021communicative}
S.~Mai, S.~Zheng, Y.~Yang, and H.~Hu, ``Communicative message passing for
  inductive relation reasoning,'' \emph{Association for the Advancement of
  Artificial Intelligence (AAAI)}, 2021.

\bibitem{WANG2021259}
S.~Wang, K.~Fu, X.~Sun, Z.~Zhang, S.~Li, and L.~Jin, ``Hierarchical-aware
  relation rotational knowledge graph embedding for link prediction,''
  \emph{Neurocomputing}, vol. 458, pp. 259--270, 2021.

\bibitem{ZHANG2022109}
Q.~Zhang, R.~Wang, J.~Yang, and L.~Xue, ``Structural context-based knowledge
  graph embedding for link prediction,'' \emph{Neurocomputing}, vol. 470, pp.
  109--120, 2022.

\bibitem{NAYYERI2021530}
M.~Nayyeri, G.~M. Cil, S.~Vahdati, F.~Osborne, M.~Rahman, S.~Angioni,
  A.~Salatino, D.~R. Recupero, N.~Vassilyeva, E.~Motta \emph{et~al.},
  ``Trans4e: Link prediction on scholarly knowledge graphs,''
  \emph{Neurocomputing}, vol. 461, pp. 530--542, 2021.

\bibitem{ShenModeling2021}
Y.~Shen, N.~Ding, H.-T. Zheng, Y.~Li, and M.~Yang, ``Modeling relation paths
  for knowledge graph completion,'' \emph{IEEE Transactions on Knowledge and
  Data Engineering}, vol.~33, no.~11, pp. 3607--3617, 2021.

\bibitem{TransE}
A.~Bordes, N.~Usunier, A.~Garcia-Duran, J.~Weston, and O.~Yakhnenko,
  ``Translating embeddings for modeling multi-relational data,'' \emph{Advances
  in neural information processing systems}, vol.~26, 2013.

\bibitem{Distmult}
B.~Yang, W.~tau Yih, X.~He, J.~Gao, and L.~Deng, ``Embedding entities and
  relations for learning and inference in knowledge bases,'' 2015.

\bibitem{ComplEx}
T.~Trouillon, C.~R. Dance, {{\'E}}ric Gaussier, J.~Welbl, S.~Riedel, and
  G.~Bouchard, ``Knowledge graph completion via complex tensor factorization,''
  \emph{Journal of Machine Learning Research}, vol.~18, no. 130, pp. 1--38,
  2017.

\bibitem{ConvE}
T.~Dettmers, P.~Minervini, P.~Stenetorp, and S.~Riedel, ``Convolutional 2d
  knowledge graph embeddings,'' in \emph{Proceedings of the AAAI Conference on
  Artificial Intelligence}, vol.~32, no.~1, 2018.

\bibitem{SACN}
C.~Shang, Y.~Tang, J.~Huang, J.~Bi, X.~He, and B.~Zhou, ``End-to-end
  structure-aware convolutional networks for knowledge base completion,'' in
  \emph{Proceedings of the AAAI Conference on Artificial Intelligence},
  vol.~33, no.~01, 2019, pp. 3060--3067.

\bibitem{R-GCN}
M.~Schlichtkrull, T.~N. Kipf, P.~Bloem, R.~Van Den~Berg, I.~Titov, and
  M.~Welling, ``Modeling relational data with graph convolutional networks,''
  in \emph{European semantic web conference}.\hskip 1em plus 0.5em minus
  0.4em\relax Springer, 2018, pp. 593--607.

\bibitem{nathani2019learning}
D.~Nathani, J.~Chauhan, C.~Sharma, and M.~Kaul, ``Learning attention-based
  embeddings for relation prediction in knowledge graphs,'' 2019.

\bibitem{PathCon}
H.~Wang, H.~Ren, and J.~Leskovec, ``Relational message passing for knowledge
  graph completion,'' in \emph{Proceedings of the 27th ACM SIGKDD Conference on
  Knowledge Discovery \& Data Mining}, 2021, pp. 1697--1707.

\bibitem{chen2021topology}
J.~Chen, H.~He, F.~Wu, and J.~Wang, ``Topology-aware correlations between
  relations for inductive link prediction in knowledge graphs,'' in
  \emph{Proceedings of the AAAI Conference on Artificial Intelligence},
  vol.~35, no.~7, 2021, pp. 6271--6278.

\bibitem{zheng2021subgraphaware}
S.~Zheng, S.~Mai, Y.~Sun, H.~Hu, and Y.~Yang, ``Subgraph-aware few-shot
  inductive link prediction via meta-learning,'' \emph{IEEE Transactions on
  Knowledge and Data Engineering}, pp. 1--1, 2022.

\bibitem{zhuang2019local}
C.~Zhuang, A.~L. Zhai, and D.~Yamins, ``Local aggregation for unsupervised
  learning of visual embeddings,'' in \emph{Proceedings of the IEEE/CVF
  International Conference on Computer Vision}, 2019, pp. 6002--6012.

\bibitem{he2020momentum}
K.~He, H.~Fan, Y.~Wu, S.~Xie, and R.~Girshick, ``Momentum contrast for
  unsupervised visual representation learning,'' in \emph{Proceedings of the
  IEEE/CVF conference on computer vision and pattern recognition}, 2020, pp.
  9729--9738.

\bibitem{giorgi2020declutr}
J.~Giorgi, O.~Nitski, B.~Wang, and G.~Bader, ``Declutr: Deep contrastive
  learning for unsupervised textual representations,'' \emph{arXiv preprint
  arXiv:2006.03659}, 2020.

\bibitem{wu2020clear}
Z.~Wu, S.~Wang, J.~Gu, M.~Khabsa, F.~Sun, and H.~Ma, ``Clear: Contrastive
  learning for sentence representation,'' \emph{arXiv preprint
  arXiv:2012.15466}, 2020.

\bibitem{you2020graph}
Y.~You, T.~Chen, Y.~Sui, T.~Chen, Z.~Wang, and Y.~Shen, ``Graph contrastive
  learning with augmentations,'' \emph{Advances in Neural Information
  Processing Systems}, vol.~33, pp. 5812--5823, 2020.

\bibitem{pennington2014glove}
J.~Pennington, R.~Socher, and C.~D. Manning, ``Glove: Global vectors for word
  representation,'' in \emph{Proceedings of the 2014 conference on empirical
  methods in natural language processing (EMNLP)}, 2014, pp. 1532--1543.

\bibitem{t-SNE}
L.~van~der Maaten and G.~Hinton, ``Viualizing data using t-sne,'' \emph{Journal
  of Machine Learning Research}, vol.~9, pp. 2579--2605, 11 2008.

\bibitem{FB15k-237}
K.~Toutanova, D.~Chen, P.~Pantel, H.~Poon, P.~Choudhury, and M.~Gamon,
  ``Representing text for joint embedding of text and knowledge bases,'' in
  \emph{Proceedings of the 2015 conference on empirical methods in natural
  language processing}, 2015, pp. 1499--1509.

\bibitem{nell}
C.~Xiong, R.~Power, and J.~Callan, ``Explicit semantic ranking for academic
  search via knowledge graph embedding,'' in \emph{Proceedings of the 26th
  international conference on world wide web}, 2017, pp. 1271--1279.

\bibitem{graph_attention}
P.~Veličković, G.~Cucurull, A.~Casanova, A.~Romero, P.~Liò, and Y.~Bengio,
  ``Graph attention networks,'' in \emph{International Conference on Learning
  Representations}, 2018.

\end{thebibliography}

%\end{thebibliography}

\end{document}